\documentclass{article}

\usepackage[english]{babel}

\usepackage[letterpaper,top=2cm,bottom=2cm,left=2cm,right=2cm,marginparwidth=1.75cm]{geometry}
\usepackage[utf8]{inputenc}
\usepackage[T1]{fontenc}
\usepackage{amsmath}
\usepackage{amsfonts}
\usepackage{amssymb}
\usepackage{bm}
\usepackage{graphicx}
\usepackage{booktabs}
\usepackage{multirow}
\usepackage{caption}
\usepackage[ruled,vlined]{algorithm2e}
\usepackage{notoccite}
\usepackage[colorlinks=true, allcolors=blue]{hyperref}
\providecommand{\keywords}[1]
{
  \small	
  \textbf{Keywords:} #1
}
\title{Dynamic Adaptive Shared Experts with Grouped Multi-Head Attention Mixture of Experts}
\author{
  Cheng Li$^{1}$, Jiexiong Liu$^{1}$, Yixuan Chen$^{1}$, Jie Ji$^{1}$\\[2pt]
  $^{1}$KunLun Meta
}

\begin{document}
\maketitle
\captionsetup[figure]{labelfont=bf, labelformat=default, labelsep=period}

\begin{abstract}
Transformer models based on the Mixture of Experts (MoE) architecture have made significant progress in long-sequence modeling, but existing models still have shortcomings in computational efficiency and the ability to capture long-range dependencies, especially in terms of the dynamic adaptability of expert resource allocation. In this paper, we propose a Dynamic Adaptive Shared Expert and Grouped Multi-Head Attention Hybrid Model (DASG-MoE) to enhance long-sequence modeling capabilities by integrating three modules. First, we employ the Grouped Multi-Head Attention (GMHA) mechanism to effectively reduce the computational complexity of long sequences. By parallel processing through sequence grouping, local sliding window attention, and feature aggregation, we address long-range dependency issues and the model's lack of generalization for local information. Second, we design a Dual-Scale Shared Expert Structure (DSSE), where shallow experts use lightweight computations to quickly respond to low-dimensional features, while deep experts process high-dimensional complex semantics through pre-training transfer and post-training optimization, achieving a dynamic balance between efficiency and accuracy. Third, we propose a hierarchical Adaptive Dynamic Routing (ADR) mechanism that dynamically selects expert levels based on feature complexity and task requirements, and optimizes resource allocation through a local expert activation strategy. Experiments on multiple long-sequence benchmark datasets demonstrate that our DASG-MoE model outperforms state-of-the-art models.
\end{abstract}
\keywords{MoE, Transformer, Grouped Multi-Head Attention, Dual-Scale Shared Expert Structure, Adaptive Dynamic Routing}
\section{Introduction}

Recent years have witnessed extraordinary advances in Natural Language Processing (NLP), driven primarily by the availability of large-scale datasets\cite{r56} and the development of sophisticated deep neural architectures, especially Transformer-based models\cite{r1}. Contemporary research demonstrates that increasing neural network size typically improves sample efficiency\cite{r9} while delivering superior generalization capabilities and enhanced predictive accuracy\cite{r3}. A notable strategy for expanding Transformer model capacity involves Mixture-of-Experts (MoE) architectures\cite{r5}, which leverage multiple specialized subnetworks and sparse activation patterns to effectively scale models to trillion-parameter regimes\cite{r6}.

\begin{figure}[htbp]
   \centering
   \includegraphics[width=0.8\textwidth]{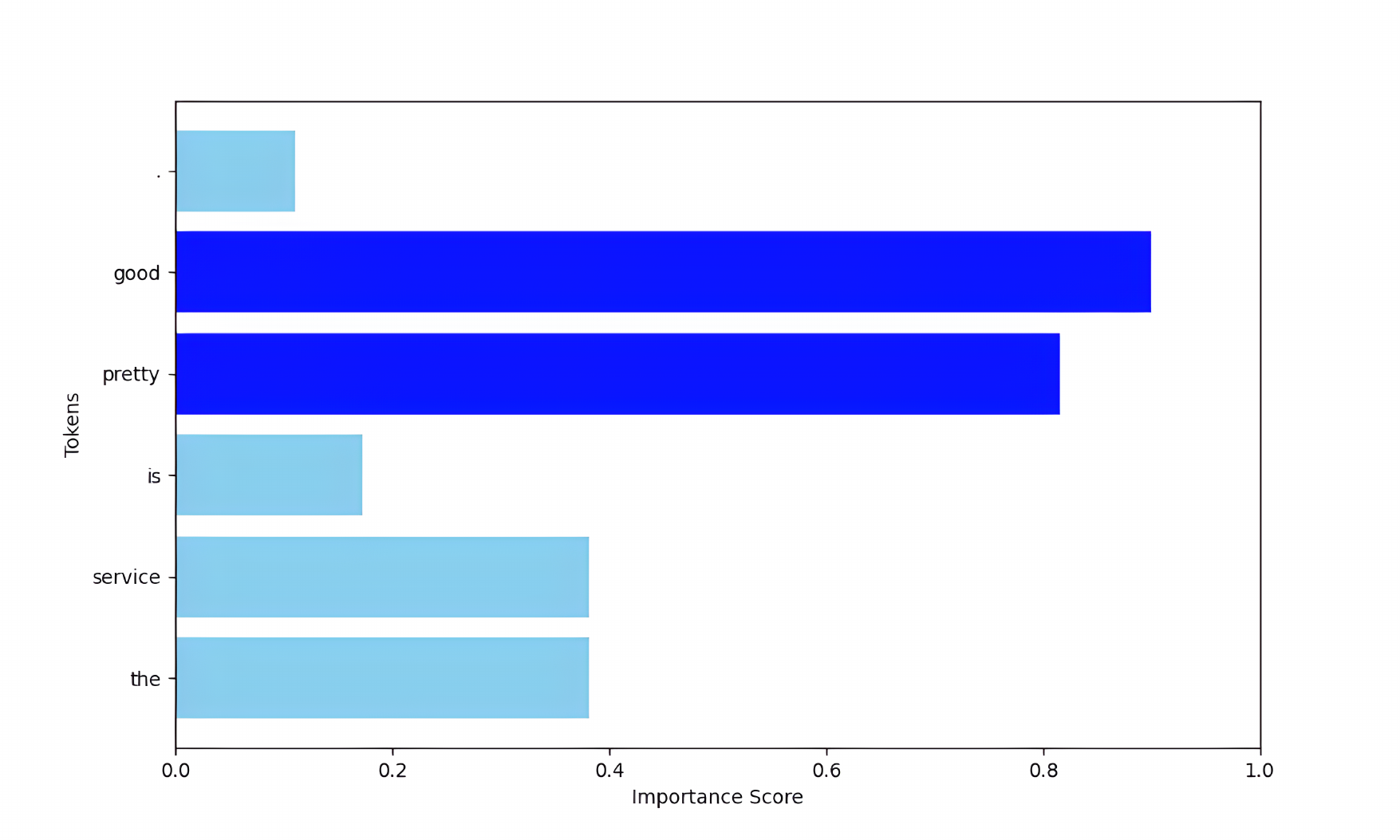}
   \caption{An example of a sentiment analysis task. }
   \label{fig:1}
\end{figure}

In MoE frameworks, a routing mechanism selects appropriate experts for processing individual tokens\cite{r21}. However, current MoE approaches exhibit a significant limitation: they assign a predetermined number of expert networks to each input token, failing to accommodate the variable significance of different tokens within the input sequence. Consider the sentiment analysis example shown in \autoref{fig:1}, where the input sentence "The service is pretty good" contains tokens of varying semantic importance. Words such as "pretty" and "good" possess considerable semantic significance\cite{r37} and are crucial for accurate classification, necessitating activation of multiple experts to effectively capture their underlying meanings. Conversely, tokens with limited semantic content such as "The" and "is" could efficiently utilize fewer experts, thereby improving computational efficiency across the entire model.

We begin with a concise overview of Transformers, MoE architectures, and their combined implementation. The Transformer architecture constitutes a prevalent deep learning framework that employs self-attention mechanisms\cite{r36}, allowing models to evaluate token significance within input sequences, model long-distance relationships, and concentrate on critical segments to effectively extract contextual information and inter-token dependencies. This architectural design demonstrates exceptional performance when handling sequential data, especially for natural language processing applications. Consequently, leading language models including BERT\cite{r2}, GPT-3\cite{r3}, ChatGPT\cite{r18}, and Llama\cite{r4} have incorporated Transformer architectures.
MoE represents a sparsely-activated network architecture\cite{r7} where routing mechanisms select specific expert subnetworks for input processing rather than engaging all available parameters, thus enhancing computational efficiency and model scalability. Integrating MoE methodologies with Transformer frameworks has demonstrated initial success in expanding model capacity to trillion-parameter scales. However, current approaches maintain static expert allocation across input tokens\cite{r30} regardless of their relative significance, creating an ongoing research challenge: developing sophisticated MoE architectures that incorporate token importance measures\cite{r24} to advance both predictive accuracy and computational efficiency in large-scale Transformer models.

Within Transformer architectures, attention layers compute attention scores by evaluating token relevance through query, key, and value vector operations, enabling models to prioritize contextually significant tokens. This inherent attention mechanism offers a promising foundation for measuring token importance, which could inform MoE routing decisions regarding expert quantity and selection\cite{r35}, ultimately enabling adaptive expert allocation strategies.

To address these challenges, we introduce a novel hybrid expert architecture: the Dynamic Adaptive Shared Expert and Group Multi-Head Attention Hybrid Model (DASG-MoE). Our proposed framework employs group multi-head attention mechanisms to process input sequences, subsequently directing them through adaptive routing strategies to specialized expert modules for computational processing. The architecture incorporates dynamic expert selection mechanisms within hybrid expert frameworks, yielding substantial improvements in both predictive accuracy and computational efficiency.
This research contributes four key innovations to the field:
\begin{itemize}
\item We developed the Dynamic Adaptive Shared Experts with Grouped Multi-Head Attention Mixture of Experts (DASG-MoE) as an innovative solution to address the limitations of conventional MoE architectures. Through systematic examination of existing MoE frameworks, we identified critical constraints, particularly their reliance on static expert allocation strategies that assign fixed experts to each input token, which limits their adaptability and efficiency in dynamic computational scenarios.
\item We approach employs a grouped multi-head attention architecture for input sequence processing, wherein sequences are partitioned into segments and processed through sliding window-based multi-head attention computations. The resulting attention weights from each segment are subsequently forwarded to dedicated MLP that perform feature transformation and consolidation operations.
\item We propose a dual-scale shared expert structure that uses dynamic routing to automatically select expert modules for input features. Shallow experts are responsible for capturing low-dimensional features, while deep experts focus on modeling complex high-dimensional semantic information.
\item We propose a method based on dynamic routing forwarding, using sequence feature vectors derived from multi-head attention outputs. By calculating feature complexity scores and urgency metrics using a lightweight evaluator, the system can automatically switch to expert mode based on the complexity of the input sequence.
\end{itemize}
\section{Related Work}
The advent of self-attention mechanisms has enabled Transformer architectures to fundamentally transform how sequence data is processed, spurring extensive research into improving their computational efficiency, scalability, and versatility for diverse practical applications. Research efforts to optimize Transformer designs can be organized into three primary categories.
The first research direction addresses the computational burden inherent in attention mechanisms, specifically targeting the quadratic scaling complexity that increases with input sequence length. Representative approaches include Linformer, Reformer, and Performer, which modify attention computation strategies to achieve reduced computational overhead while preserving the fundamental advantages of self-attention architectures.The second research stream concentrates on parameter reduction strategies that maintain model performance while decreasing memory requirements. ALBERT exemplifies this approach by implementing cross-layer parameter sharing techniques, achieving significant model compression without sacrificing predictive accuracy.The third research category encompasses MoE methodologies, which have demonstrated remarkable success in scaling Transformer models to trillion-parameter sizes while avoiding proportional computational cost increases. These techniques enable efficient model scaling through selective expert activation.

This research advances the third category by introducing a novel Dynamic Adaptive Shared Expert and Grouped Multi-Head Attention Hybrid Model (DASG-MoE) architecture that incorporates adaptive expert selection mechanisms.

In the domain of deep learning, the MoE architecture has propelled numerous innovations, especially regarding the expansion of models designed for natural language processing. Initially introduced by Jacobs and colleagues during the 1990s, this approach sought to develop specialized networks that concentrate on distinct portions of the data domain. Fundamental to MoE systems is a gating network, often termed a router, tasked with identifying the most suitable specialist for handling specific instances. Subsequent portions of this discussion outline key strategies for routing within these frameworks:
\begin{itemize}
\item The Switch Transformer\cite{r6}, introduced by Fedus and collaborators at Google, streamlines the gating mechanism in MoE architectures through assigning individual tokens solely to one specialist for computation. While this configuration substantially decreases data transfer overhead and bolsters the robustness of training procedures, it continues to rely on static selection protocols for experts, without provisions for varying the quantity of active specialists in response to differing levels of input sophistication.
\item Google's GLaM\cite{r7} architecture incorporates a mechanism for selecting the top two experts, directing individual tokens to those exhibiting the peak activation levels. Relative to conventional dense networks, this strategy considerably enhances both parametric utilization and overall scale, yet the invariant top-k policy precludes adaptive variation in the volume of engaged specialists contingent upon input challenges or intricate attributes.
\item Within the PaLM-2\cite{r8} framework, the MoE element applies a trainable gating structure to manage specialist assignments, facilitating a weighted integration across various specialists via smooth activation techniques. Nonetheless, its training regimen necessitates advance determination of both the aggregate specialist pool for operations and the ceiling on activations per input element, which inherently curtails the system's capacity for flexible adaptation to varying degrees of input intricacy.
\end{itemize}

\section{Problem Statement}
Transformer-based MoE models for long-sequence inputs are hampered by two main issues: static expert allocation and quadratic attention cost\cite{r38}. Consider an input token sequence $X={x_1,\dots,x_N}$ of length $N$. Current MoE routers typically assign each token $x_i$ a constant number of experts $K$, regardless of its importance:
\begin{equation}
    K_i = K
\end{equation}
where $K_i$ is the number of experts for token $i$. This static allocation forces both high-importance and low-importance tokens to use equal resources, wasting computation on trivial tokens\cite{r10}. At the same time, standard self-attention over $N$ tokens scales as $O(N^2)$
, which is prohibitive for large $N$ and limits the model’s ability to capture long-range dependencies\cite{r14,r39}. In summary, fixed-$K$ routing and naïve full attention make long sequences expensive and inefficient to model.

Ideally, the router should be dynamic and adapt to each token’s complexity. Let $I_i$ be an importance score for token $x_i$ (e.g.\ derived from attention activations)\cite{r28}. A dynamic routing mechanism would allocate experts based on $I_i$:
\begin{equation}
\label{eq:2}
    K_i = f(I_i)
\end{equation}
where $f(\cdot)$ is a non-decreasing function, the calculation is as follows:
\begin{equation}
    f(I_i) = \lceil L_i \times K \rceil
\end{equation}
here, $\lceil\cdot\rceil$ denotes rounding up to ensure K is an integer.
If $I_i$ is large, the complexity is higher, and K is larger (more experts). If $I_i$ is small, the complexity is lower, and K is smaller (fewer experts). In this scheme, tokens with larger $I_i$ receive more experts\cite{r32}, and tokens with smaller $I_i$ receive fewer, improving resource efficiency. However, alone does not resolve the long-sequence bottleneck: it still assumes $O(N^2)$ attention or naïve sparse routing\cite{r26} and does not exploit the structure of the input sequence.

Another limitation is how long-range versus local context are handled. Naïve sliding-window attention limits each token to a local window of width $w$ but then ignores distant tokens. To balance this, one can partition $X$ into $G$ contiguous segments $X_1,\dots,X_G$ and apply local attention within each segment. More generally, Grouped Multi-Head Attention (GMHA) reshapes queries, keys and values into $G$ groups\cite{r40}, reducing attention complexity from $O(N^2)$ to $O(Nw)$\cite{r41}, where $w$ denotes the size of the sliding window. GMHA thus enables efficient long-range modeling by grouping tokens and sliding windows, but it requires a principled way to combine global and local information\cite{r43}. Finally, conventional MoE experts are unstructured: every token is processed by the same kind of expert networks\cite{r29,r33,r34}. We introduce a Dual-Scale Shared Expert (DSSE) structure with two tiers of experts: shallow experts $\mathcal{E}s$ and deep experts $\mathcal{E}_d$\cite{r25}. Each token $x_i$ is routed to $k_{s,i}$ shallow and $k{d,i}$ deep experts (with $k_{s,i}+k_{d,i}=K_i$). A static tier allocation would be fixed as follows:
\begin{equation}
\label{eq:3}
    (k_{s,i}, k_{d,i}) = (k_s^0, k_d^0)
\end{equation}
here, giving every token the same number of shallow vs.\ deep experts. 

Instead, we pursue a hierarchical dynamic routing: choose $(k_{s,i},k_{d,i})$ based on a complexity measure $C_i$ of token $x_i$\cite{r22}. Concretely, we allow the following:
\begin{equation}
\label{eq:4}
    (k_{s,i}, k_{d,i}) = g(C_i, T_j)
\end{equation}
\begin{equation}
\label{eq:5}
    k_{s,i} + k_{d,i} = K_i,
\end{equation}
To incorporate task-adaptive routing, we extend $g$ to include a task type vector $T_j \in \mathbb{R}^2$ (e.g., $T_{MRPC} = [0.2, 0.8]$ for emphasizing deep experts in complex semantics). The updated function is:
\begin{equation}
    g(C_i, T_j) = K_i \cdot (T_j[0] + (1 - T_j[0]) \cdot (1 - C_i))
\end{equation}

where $g(\cdot)$ allocates more deep experts when $C_i$ is larger\cite{r50} and more shallow experts otherwise\cite{r51}. To optimize this and prevent over-processing of simple features by deep experts, we introduce a threshold-based function:
\begin{equation}
    g(C_i) = 
    \begin{cases} 
        (K_i, 0) & \text{if } C_i < \theta_s \\
        (0, K_i) & \text{if } C_i > \theta_d \\
        \left( \left\lfloor \frac{K_i}{2} \right\rfloor, \left\lceil \frac{K_i}{2} \right\rceil \right) & \text{otherwise}
    \end{cases}
\end{equation}
where $\theta_s = 0.3$ and $\theta_d = 0.7$ are empirically tuned thresholds to ensure balanced allocation and avoid excessive deep expert usage on low-complexity tokens, which can degrade performance in tasks like MRPC requiring semantic equivalence detection\cite{r61}.

In summary, the existing MoE-Transformer architecture for processing long sequences faces three core challenges: the quadratic cost of the attention mechanism, token routing with a fixed $K$, and undifferentiated expert networks\cite{r23,r27}. We have formalized these challenges using the above equations, which describe static versus dynamic expert allocation, sequence grouping into segments $G$ with window size $w$, and multi-tier expert selection. The proposed DASG-MoE model addresses each of these issues. Specifically, DASG-MoE replaces naïve $O(N^2)$ attention with grouped attention\cite{r15}, reducing complexity to $O(N^2/g)$\cite{r16}, and substitutes static routing with adaptive allocation strategies\cite{r17} defined in \autoref{eq:2}, \autoref{eq:3}, \autoref{eq:4} and \autoref{eq:5}. These design choices mitigate identified inefficiencies by lowering computational cost and aligning expert resources with token complexity, thereby enabling efficient modeling of extremely long contexts.

\section{DASG-MoE Overview}
We propose a Dynamic Adaptive Shared Experts with Grouped Multi-Head Attention Mixture of Experts (DASG-MoE), which utilizes Group Multi-Head Attention to process input sequences and dynamic routing to forward them to the corresponding experts.  This section outlines our DASG-MoE model, including the model architecture and algorithm description.

\subsection{DASG-MoE Architecture}
In \autoref{fig:2}, the allocation of experts to individual tokens depends on their computed scores, where tokens achieving higher scores are assigned to a greater number of expert modules. A distinctive characteristic of this architecture involves the stratified organization of expert components, employing deeper expert layers for processing complex, high-dimensional feature representations while utilizing shallower expert networks to handle simpler, low-dimensional characteristics. Following the computation of individual token scores, the evaluation mechanism determines a complexity metric for each token and routes it adaptively to the appropriate expert module based on this assessment.Key components of our DASG-MoE approach include:
\begin{figure}[htbp]
   \centering
   \includegraphics[width=0.8\textwidth]{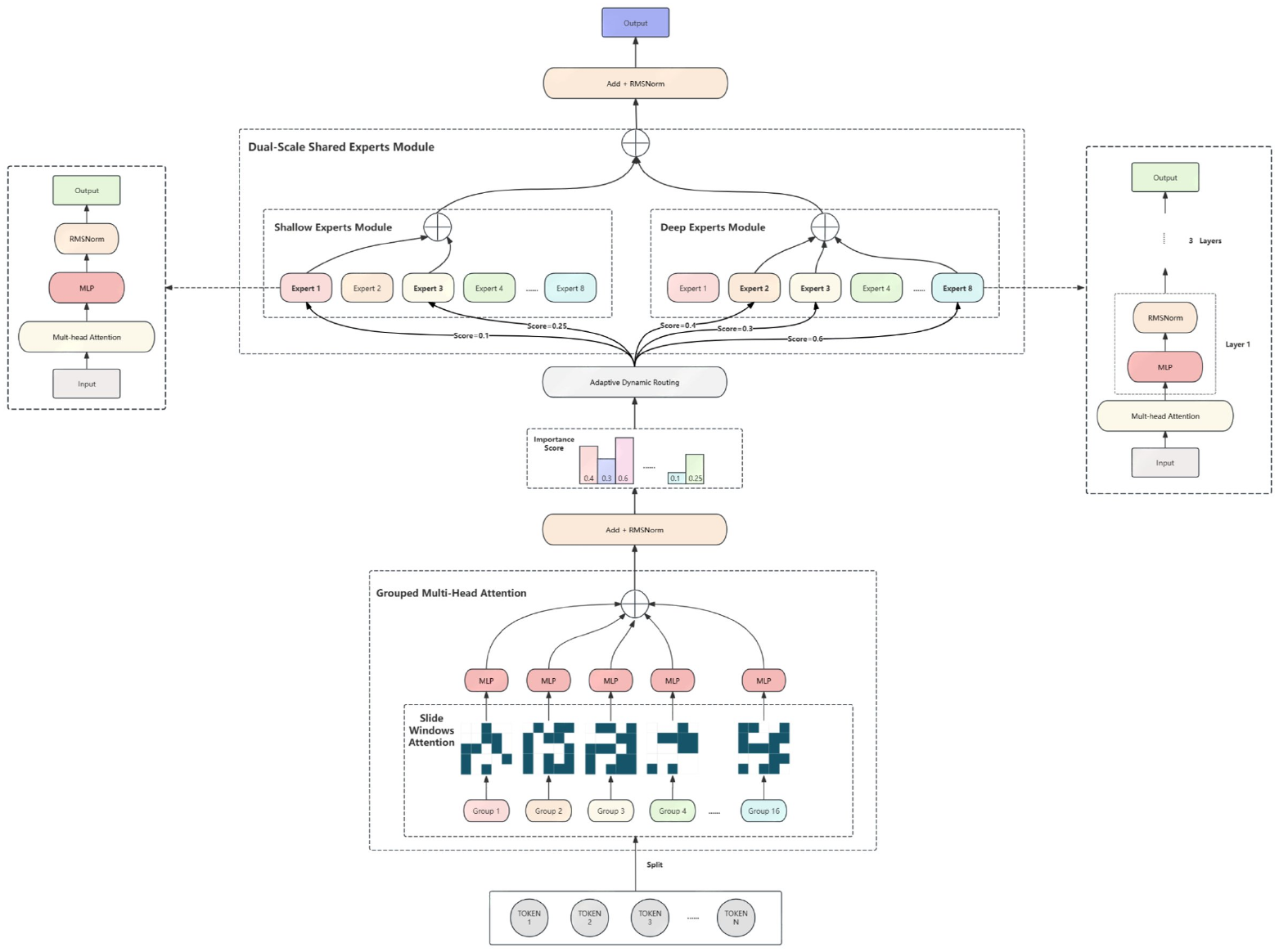}
   \caption{Our adaptive routing mechanism and MoE system diagram of the dual-scale shared expert module allow the model to assign corresponding hierarchical experts based on token importance. }
   \label{fig:2}
\end{figure}

\begin{itemize}
    \item \textbf{Grouped Multi-Head Attention}: For the input sequence, the sequence is divided into 16 groups, and the multi-head attention mechanism with a sliding window is applied in parallel to each group to calculate the attention. Then, the attention results of each group are transformed using MLP, and finally, all the calculation results are merged.
    \item \textbf{Dual-Scale Shared Experts}: Set up two expert modules, with the shallow expert module consisting of 8 experts, each equipped with a single-layer MLP, and the deep expert module consisting of 8 experts, each equipped with a three-layer MLP. The deep expert module is obtained by replicating the pre-trained shallow expert module, with two additional layers of MLP added after the first layer, each having the same structure as the first layer. During training, the first layer of the MLP is frozen, and fine-tuning is performed using a dataset for complex tasks to acquire the ability to handle complex tasks. The global router is trained using a time reward function and an accuracy reward function to select either the shallow or deep expert module based on the task type. Within each expert module, the local router controls the selection of the top 2 experts for computation.
    \item \textbf{Adaptive Dynamic Routing}: First, the feature vectors of the input sequence are extracted through a grouped multi-head attention mechanism, and a lightweight evaluator is used to calculate the feature complexity score and task urgency index, providing quantitative basis for subsequent routing decisions. Then, based on these evaluation metrics, the global router selects the appropriate level in the dual-scale expert architecture by combining time and accuracy dual reward functions.
\end{itemize}

\subsection{Grouped Multi-Head Attention}
We propose an innovative architecture based on a grouped multi-head attention mechanism. This architecture divides the input sequence $X\in \mathbb{R}^{N \times d}$ uniformly into $G=16$ subgroups $\{X_1, X_2, ..., X_{16}\}$ along the sequence dimension, Each subgroup $X_i \in \mathbb{R}^{(N/G) \times d}$ Specifically, for the $i$-th subgroup, the model first applies a multi-head attention mechanism with a sliding window for feature extraction, with the calculation formula being: 
\begin{equation}
    \text{Attention}_i(Q_i, K_i, V_i) = \text{Softmax}\left(\frac{Q_i K_i^T}{\sqrt{d_k}} + M_{\text{window}}\right)V_i
\end{equation}
, where $Q_i$, $K_i$, $V_i$ represent the query, key, and value matrices for the i-th group, respectively, $d_k$ is the key dimension, and $M_{\text{window}} \in \mathbb{R}^{(N/G) \times (N/G)}$ is the sliding window mask matrix, which only allows local attention calculations with a window size of w, the calculation formula is as follows,

\begin{equation}
    M_{\text{window}}[i,j] = \begin{cases}
0, & \text{if } {|i - j| \le \lfloor w/2 \rfloor} \\
-\infty, & \text{if } {|i - j| > \lfloor w/2 \rfloor}
\end{cases}
\end{equation}

Subsequently, the attention output $H_i$ of each subgroup undergoes parallel feature refinement through an independent multi-layer perceptron, i.e :

\begin{equation}
    \text{MLP}_i(H_i) = W_{2i} \cdot \text{ReLU}(W_{1i} \cdot H_i + b_{1i}) + b_{2i}
\end{equation}
, where $W_{1i} \in \mathbb{R}^{d_{\text{ff}} \times d}$, $W_{2i} \in \mathbb{R}^{d \times d_{\text{ff}}}$ is the weight matrix specific to the $i$-th group, and $b_{1i}$, $b_{2i}$ are the corresponding bias terms. Finally, the results from all groups are fused through a feature aggregation operation: 

\begin{equation}
    \text{Output} = \text{Concat}(\text{MLP}_1(H_1), \text{MLP}_2(H_2), \ldots, \text{MLP}_{16}(H_{16})) \cdot W_{\text{out}} + b_{\text{out}}
\end{equation}
, where $W_{\text{out}} \in \mathbb{R}^{d \times d}$ is the output projection matrix.

The core advantages of this design are: (1) significantly improved computational efficiency and reduced overall complexity through parallel computing; (2) group processing is naturally suited to long sequence modelling, with information exchange between different groups achieved through the final aggregation layer, effectively alleviating the computational bottleneck of long-range dependency modelling; (3) The sliding window attention mechanism reinforces local feature learning, improving the model's ability to capture local patterns in sequences and its generalisation performance, while also enhancing the robustness of the overall architecture.

\subsection{Dual-Scale Shared Expert Network Algorithm}
We introduce the proposed Dual-Scale Shared Expert Network (DSSEN) approach for efficient mixture-of-experts inference. \autoref{alg:1} presents a sketch of the key steps, including (1) global routing decision to select between shallow and deep expert modules based on input characteristics, (2) local expert selection using top-k gating within the chosen module, (3) forward computation through either 1-layer shallow experts or 3-layer deep experts with residual connections, and (4) weighted combination of selected expert outputs with final projection. Details are as follows:

\begin{algorithm}[H]
\caption{Dual-Scale Shared Expert Network with Dynamic Routing}
\label{alg:1}

\KwIn{Input feature $\mathbf{x} \in \mathbb{R}^{d_{in}}$}
\KwOut{Network output $\mathbf{y} \in \mathbb{R}^{d_{out}}$}

\Begin{
    \textbf{Step 1: Global Routing Decision}\\
    Compute global routing probabilities: $\mathbf{p}_g = \text{Softmax}(R_g(\mathbf{x})) \in \mathbb{R}^2$\\
    where $R_g: \mathbb{R}^{d_{in}} \rightarrow \mathbb{R}^2$ is the global router\\
    Select expert module: $m = \arg\max(\mathbf{p}_g) \in \{0, 1\}$ (0: shallow, 1: deep)\\
    
    \textbf{Step 2: Local Expert Selection and Processing}\\
    \If{$m = 0$ (Shallow Module)}{
        Compute shallow local routing: $\mathbf{p}_s = \text{Softmax}(R_s(\mathbf{x})) \in \mathbb{R}^8$\\
        where $R_s: \mathbb{R}^{d_{in}} \rightarrow \mathbb{R}^8$ is the shallow local router\\
        
        Select top-k experts: $\{i_1, i_2\} = \text{top-k}(\mathbf{p}_s,k=2)$\\
        
        \For{$k \in \{i_1, i_2\}$}{
            Compute shallow expert output: $\mathbf{h}_s^{(k)} = \sigma(\mathbf{W}_s^{(k)}\mathbf{x} + \mathbf{b}_s^{(k)})$\\
            where $\mathbf{W}_s^{(k)} \in \mathbb{R}^{d_{hidden} \times d_{in}}$, $\mathbf{b}_s^{(k)} \in \mathbb{R}^{d_{hidden}}$\\
        }
        
        Weighted combination: $\mathbf{y} = p_s^{(i_1)} \mathbf{h}_s^{(i_1)} + p_s^{(i_2)} \mathbf{h}_s^{(i_2)}$\\
        Final output projection: $\mathbf{y} = \mathbf{W}_{out}^s \mathbf{y} + \mathbf{b}_{out}^s$\\
        where $\mathbf{W}_{out}^s \in \mathbb{R}^{d_{out} \times d_{hidden}}$, $\mathbf{b}_{out}^s \in \mathbb{R}^{d_{out}}$\\
    }
    \Else{
        Compute deep local routing: $\mathbf{p}_d = \text{Softmax}(R_d(\mathbf{x})) \in \mathbb{R}^8$\\
        where $R_d: \mathbb{R}^{d_{in}} \rightarrow \mathbb{R}^8$ is the deep local router\\
        
        Select top-k experts: $\{j_1, j_2\} = \text{top-k}(\mathbf{p}_d,k=2)$\\
        
        \For{$k \in \{j_1, j_2\}$}{
            \textbf{Layer 1 (frozen):} $\mathbf{h}_d^{(k,1)} = \sigma(\mathbf{W}_d^{(k,1)}\mathbf{x} + \mathbf{b}_d^{(k,1)})$\\
            where $\mathbf{W}_d^{(k,1)} \in \mathbb{R}^{d_{hidden} \times d_{in}}$, $\mathbf{b}_d^{(k,1)} \in \mathbb{R}^{d_{hidden}}$\\
            
            \textbf{Layer 2:} $\mathbf{h}_d^{(k,2)} = \sigma(\mathbf{W}_d^{(k,2)}\mathbf{h}_d^{(k,1)} + \mathbf{b}_d^{(k,2)})$\\
            where $\mathbf{W}_d^{(k,2)} \in \mathbb{R}^{d_{hidden} \times d_{hidden}}$, $\mathbf{b}_d^{(k,2)} \in \mathbb{R}^{d_{hidden}}$\\
            
            \textbf{Layer 3 with residual:} $\mathbf{h}_d^{(k,3)} = \sigma(\mathbf{W}_d^{(k,3)}\mathbf{h}_d^{(k,2)} + \mathbf{b}_d^{(k,3)}) + \mathbf{h}_d^{(k,1)}$\\
            where $\mathbf{W}_d^{(k,3)} \in \mathbb{R}^{d_{hidden} \times d_{hidden}}$, $\mathbf{b}_d^{(k,3)} \in \mathbb{R}^{d_{hidden}}$\\
        }
        
        Weighted combination: $\mathbf{y} = p_d^{(j_1)} \mathbf{h}_d^{(j_1,3)} + p_d^{(j_2)} \mathbf{h}_d^{(j_2,3)}$\\
        Final output projection: $\mathbf{y} = \mathbf{W}_{out}^d \mathbf{y} + \mathbf{b}_{out}^d$\\
        where $\mathbf{W}_{out}^d \in \mathbb{R}^{d_{out} \times d_{hidden}}$, $\mathbf{b}_{out}^d \in \mathbb{R}^{d_{out}}$\\
    }
    
    \Return{$\mathbf{y} \in \mathbb{R}^{d_{out}}$}
}

\end{algorithm}

\begin{itemize}
    \item \textbf{Global Routing Decision}: The first step involves determining the appropriate expert module based on input complexity through a global router. The global router $R_g$ analyzes the input feature $X \in \mathbb{R}^{din}$ and produces routing probabilities for both shallow and deep expert modules. This binary decision mechanism is trained using reinforcement learning with combined time and accuracy rewards to automatically balance computational efficiency and task performance. The router learns to assign simple tasks to shallow experts for faster inference while directing complex tasks to deep experts for higher accuracy. The calculation process is as follows:
    \begin{equation}
        P_g = \text{Softmax}(R_g(x))
    \end{equation}
    \begin{equation}
        m = \arg\max(P_g)
    \end{equation}
    \item \textbf{Local Expert Selection via Top-k Gating}: Within the selected expert module (shallow or deep), a local router determines which specific experts to activate using a top-k gating mechanism. The local router $R_s$ computes attention weights across all 8 experts within the chosen module, where each expert specializes in different aspects of the feature space. The top-k selection ensures load balancing while maintaining computational efficiency, as only the two most relevant experts are activated for each input. This sparse activation reduces computational overhead compared to dense expert utilization while preserving model expressiveness. The calculation process is as follows:
    \begin{equation}
        p_s = \text{Softmax}(R_s(x))
    \end{equation}
    \begin{equation}
        {i_1, i_2} = \text{top-k}(p_s,k=2)
    \end{equation}
    \item \textbf{Forward Computation Through Expert Architectures}: The core computation differs significantly between shallow and deep expert modules to accommodate varying task complexities. Shallow experts employ single-layer MLPs with weight matrices $\mathbf{W}_s^{(k)} \in \mathbb{R}^{d_{hidden} \times d_{in}}$ for rapid feature transformation, suitable for tasks requiring quick responses. Deep experts utilize three-layer architectures where the first layer is copied from pre-trained shallow experts and frozen during training, while the additional layers $\mathbf{W}_d^{(k,2)}$ and $\mathbf{W}_d^{(k,3)} \in \mathbb{R}^{d_{hidden} \times d_{hidden}}$ are trained on complex tasks. Residual connections between the first and third layers facilitate gradient flow and enhance representational capacity. The calculation process is as follows:
    \begin{equation}
        h_d^{(k,3)} = \sigma(W_d^{(k,3)} h_d^{(k,2)} + b_d^{(k,3)}) + h_d^{(k,1)}
    \end{equation}
    \item \textbf{Weighted Combination and Output Projection}: The final step aggregates outputs from the selected top-k experts using their respective routing probabilities as weights, followed by a learned projection to the desired output dimensionality. The weighted combination leverages the routing probabilities $p_s^{(i)}$ or $p_d^{(j)}$ computed during local expert selection to create a smooth interpolation between expert predictions. The output projection matrices $\mathbf{W}_{out}^s$, $\mathbf{W}_{out}^d \in \mathbb{R}^{d_{out} \times d_{hidden}}$ are module-specific, allowing for different output transformations optimized for shallow versus deep expert representations. This design ensures that the final output $\mathbf{y} \in \mathbb{R}^{d_{out}}$ effectively combines expert knowledge while maintaining the desired output format. The calculation process is as follows:
    \begin{equation}
        y = W_{out} \left( \sum_{k \in top-k,k=2} p^{(k)} h^{(k)} \right) + b_{out}
    \end{equation}
\end{itemize}

\subsection{Lightweight Evaluator}

We have constructed a lightweight evaluation module. This module adopts a two-layer multi-layer perceptron architecture with a hidden layer dimension of $h$. The evaluator takes the internally generated attention weight distribution of the model as input, aiming to map it onto a two-dimensional metric space. This yields two outputs: the first is $I_i$, quantifying the contribution of this attention pattern to the model's final prediction; the second is $C_i$, assessing the intrinsic structural characteristics of the attention distribution. The calculation formula is as follows:
\begin{equation}
    [I_i, C_i]^T = W_2 \cdot \text{ReLU}(W_1 \cdot \text{AttnWeights}(x_i) + b_1) + b_2
\end{equation}
here, $AttnWeights(x_i)$ denotes the attention weight vector generated by the attention mechanism within the model after processing the input $x_i$.

Taking the sentiment analysis in \autoref{fig:1} as an example, the $I_i$=0.82 of 'pretty' triggers 3 experts, while the $I_i$=0.11 of 'the' only activates 1 expert, which verifies the effectiveness of our evaluator.

\section{Experimental Analysis}
For our experimental validation, we incorporate the proposed DASG-MoE framework within Switch Transformer, a leading Mixture-of-Experts Transformer architecture that functions as our comparative baseline. Our comprehensive evaluation methodology encompasses three distinct experimental phases: (1) initial pre-training assessment utilizing 5T tokens obtained from the Internet, (2) subsequent fine-tuning validation employing pre-trained model configurations on the C4 (Colossal Clean Crawled Corpus) dataset\cite{r12,r57}, and (3) comprehensive token significance analysis\cite{r11} spanning diverse natural language processing applications.
To ensure robust performance assessment across multiple NLP domains following fine-tuning procedures, we employ the General Language Understanding Evaluation (GLUE) benchmark  according to its established protocols\cite{r54,r55}. Our experimental infrastructure consists of 8 NVIDIA A100 GPU units, providing the computational resources necessary for thorough model evaluation and comparison.

\subsection{Experimental Setup}
The methodological framework and experimental parameters are comprehensively documented below.
\begin{enumerate}
    \item \textbf{Benchmarks}: MiniCPM4 is primarily pre-trained using Chinese and English corpora. Therefore, we selected the following datasets to evaluate model performance, specifically the knowledge-intensive evaluation set MMLU\cite{r65}, CMMLU\cite{r66}, and CEval\cite{r67}, which are used for English and Chinese evaluations, respectively. Additionally, we include reasoning evaluation datasets, including the general reasoning dataset BigBench Hard (BBH)\cite{r68}, mathematical reasoning GSM8K\cite{r69} and MATH500 \cite{r70}, as well as code reasoning MBPP\cite{r71} and HumanEval \cite{r72}.
    
    \item \textbf{Baseline Models}: We compared DASG-MoE-Instruct against several widely adopted open-source large language models, including Qwen3-8B\cite{r73}, GLM4-9B\cite{r74}, Gemma3-12B\cite{r75}, LLaMA3.1-8B\cite{r76}, and Phi4-14B\cite{r77}. The implementation results are shown in \autoref{table1}. Among these, DASG-MoE-Instruct is obtained by fine-tuning the DASG-MoE model with instructions.
    
    \item \textbf{Pre-training Settings}: The pre-training process for the DASG-MoE and Switch Transformer models utilized 5T tokens obtained from the Internet. The model architecture employs an expert configuration, including 8, 16, and 32 units\cite{r8}, with shallow modules containing 8 experts and deep modules containing 8 experts also included in the calculation\cite{r49}. It features 768 dimensional token embeddings, 3072 dimensional hidden layer processing, and 12 attention heads.
    \item \textbf{Fine-tuning Settings}: Our experimental setup is based on the pre-trained ST-Base model parameters, which uses 16 units configured by experts. In order to establish the DASG-MoE and ST architectures for subsequent fine-tuning processes, we replaced the original experts with 8 deep experts and 8 shallow experts, enabling comprehensive evaluation on the GLUE benchmark suite.

    \hspace*{2em}The General Language Understanding Evaluation benchmark encompasses diverse natural language processing challenges organized into distinct categories. Single-sentence classification encompasses SST-2\cite{r59}, which performs binary sentiment analysis on movie review data to determine positive or negative polarity, alongside CoLA\cite{r60}, designed for assessing linguistic acceptability and grammatical comprehension.

    \hspace*{2em}Similarity and paraphrase identification tasks include MRPC\cite{r61} (Microsoft Research Paraphrase Corpus) for detecting semantic equivalence, and QQP (Quora Question Pairs) for identifying duplicate question instances. Additionally, the benchmark incorporates four natural language inference challenges: MNLI\cite{r62} (Multi-Sentence Natural Language Inference) for cross-genre textual entailment, QNLI\cite{r63} (Question Natural Language Inference) for question-answering inference, RTE\cite{r64} (Recognizing Textual Entailment) for premise-hypothesis relationships, and WNLI (Winograd Natural Language Inference) for pronoun resolution tasks.
\end{enumerate}

\begin{table}[]
\caption{Evaluation results of DAGS-MoE-Instruct and other open-source LLMs.} 
\label{table1}
\centering
\begin{tabular}{ccccccc}
\hline
Models      & Qwen3          & GLM4  & Gemma3 & LLaMA3.1 & Phi4           & DASG-MoE-Instruct       \\ \hline
\#Parameter & 8B             & 9B    & 12B    & 8B       & 14B            & 8B             \\ \hline
MMLU        & 76.57          & 75.7  & 73.86  & 70.07    & 81.88          & \textbf{82.03} \\
CMMLU       & 77.23          & 74.51 & 61.24  & 55.23    & 66.93          & \textbf{78.59} \\
CEval       & 81.49          & 73.96 & 61.98  & 51.76    & 65.02          & \textbf{82.44} \\
BBH         & 68.31          & 60.8  & 65.83  & 45.34    & 73.19          & \textbf{74.1}  \\
GSM8K       & 92.14          & 90.14 & 94.18  & 84.67    & \textbf{95.13} & 92.38          \\
MATH500     & \textbf{83.89} & 66.42 & 82.3   & 49.32    & 79.77          & 78.63          \\
MBPP        & 77.18          & 74.24 & 84.52  & 68.11    & \textbf{79.89} & 76.59          \\
HumanEval   & 86.04          & 82.78 & 82.79  & 70.87    & 87.45          & \textbf{88.03} \\ \hline
Average     & 80.36          & 74.82 & 75.84  & 61.92    & 78.66          & \textbf{81.6}           \\ \hline
\end{tabular}
\end{table}

\begin{table}[]
\caption{Results for each task are highlighted in bold. B and E represent millions, billions, and experts, respectively.} \label{table2}
\centering
\begin{tabular}{@{}cccccccc@{}}
\toprule
\multirow{2}{*}{\textbf{GLUE}} & \multirow{2}{*}{\textbf{Metrics}} & \multicolumn{2}{c}{\textbf{0.6B with 8/8E}} & \multicolumn{2}{c}{\textbf{3B with 8/8E}} & \multicolumn{2}{c}{\textbf{8B with 8/8E}} \\ \cmidrule(l){3-8} 
                               &                                   & \textbf{Baseline}      & \textbf{Ours}      & \textbf{Baseline}     & \textbf{Ours}     & \textbf{Baseline}     & \textbf{Ours}     \\ \midrule
\textbf{CoLA}                  & Accuracy                          & 60.33                  & \textbf{65.53}     & 66.84                 & \textbf{67.02}    & 67.33                 & \textbf{68.83}    \\
\textbf{SST-2}                 & Accuracy                          & \textbf{78.86}                  & 77.72              & \textbf{79.9}                  & 79.81             & \textbf{80.27}                 & 80.13             \\
\textbf{MRPC}                  & Accuracy                          & \textbf{69.54}                  & 67.71              & \textbf{74.46}                 & 72.55             & \textbf{75.52}                 & 73.01             \\
\textbf{QQP}                   & Accuracy                          & 68.2                   & \textbf{69.97}     & 70.3                  & \textbf{73.29}    & 71.06                 & \textbf{75.7}     \\
\textbf{MNLI}                  & Accuracy                          & 62.76                  & \textbf{63.9}      & 68.86                 & \textbf{71.77}    & 70.97                 & \textbf{74.79}    \\
\textbf{QNLI}                  & Accuracy                          & 61.13                  & \textbf{61.96}     & 58.54                 & \textbf{61.63}    & 60.4                  & \textbf{62.22}    \\
\textbf{RTE}                   & Accuracy                          & 48.45                  & \textbf{50.52}     & 51.57                 & \textbf{51.95}    & 52.17                 & \textbf{53.4}     \\
\textbf{WNLI}                  & Accuracy                          & 46.84                  & \textbf{49.76}     & 55.63                 & \textbf{56.08}    & 56.04                 & \textbf{56.86}    \\ \midrule
\multicolumn{2}{c}{\textbf{Average}}                               & 62.01                  & \textbf{63.38}     & 65.76                 & \textbf{66.76}    & 66.72                 & \textbf{68.12}    \\ \bottomrule
\end{tabular}
\end{table}

\begin{table}[]
\caption{Results of fine-tuning using GLUE sub-tasks after pre-training on the C4 dataset. The best results for each task are highlighted in bold.} 
\label{table3}
\centering
\begin{tabular}{@{}cccc@{}}
\toprule
\multirow{2}{*}{\textbf{GLUE}} & \multirow{2}{*}{\textbf{Metrics}} & \multicolumn{2}{c}{\textbf{Our Experiments}} \\ \cmidrule(l){3-4} 
                               &                                   & \textbf{Baseline}      & \textbf{Ours}       \\ \midrule
\textbf{CoLA}                  & Accuracy                          & \textbf{70.55}         & 69.33               \\
\textbf{SST-2}                 & Accuracy                          & 95.41                  & \textbf{95.89}      \\
\textbf{MRPC}                  & Accuracy                          & 87.63                  & \textbf{88.42}      \\
\textbf{QQP}                   & Accuracy                          & 88.7                   & \textbf{89.63}      \\
\textbf{MNLI}                  & Accuracy                          & 86.88                  & \textbf{90.02}      \\
\textbf{QNLI}                  & Accuracy                          & \textbf{90.73}         & 89.4                \\
\textbf{RTE}                   & Accuracy                          & 60.29                  & \textbf{61.21}      \\
\textbf{WNLI}                  & Accuracy                          & 54.42                  & \textbf{56.98}      \\ \bottomrule
\end{tabular}
\end{table}

\subsection{Pre-training Evaluation}
\begin{figure}[htbp]
   \centering
   \includegraphics[width=0.8\textwidth]{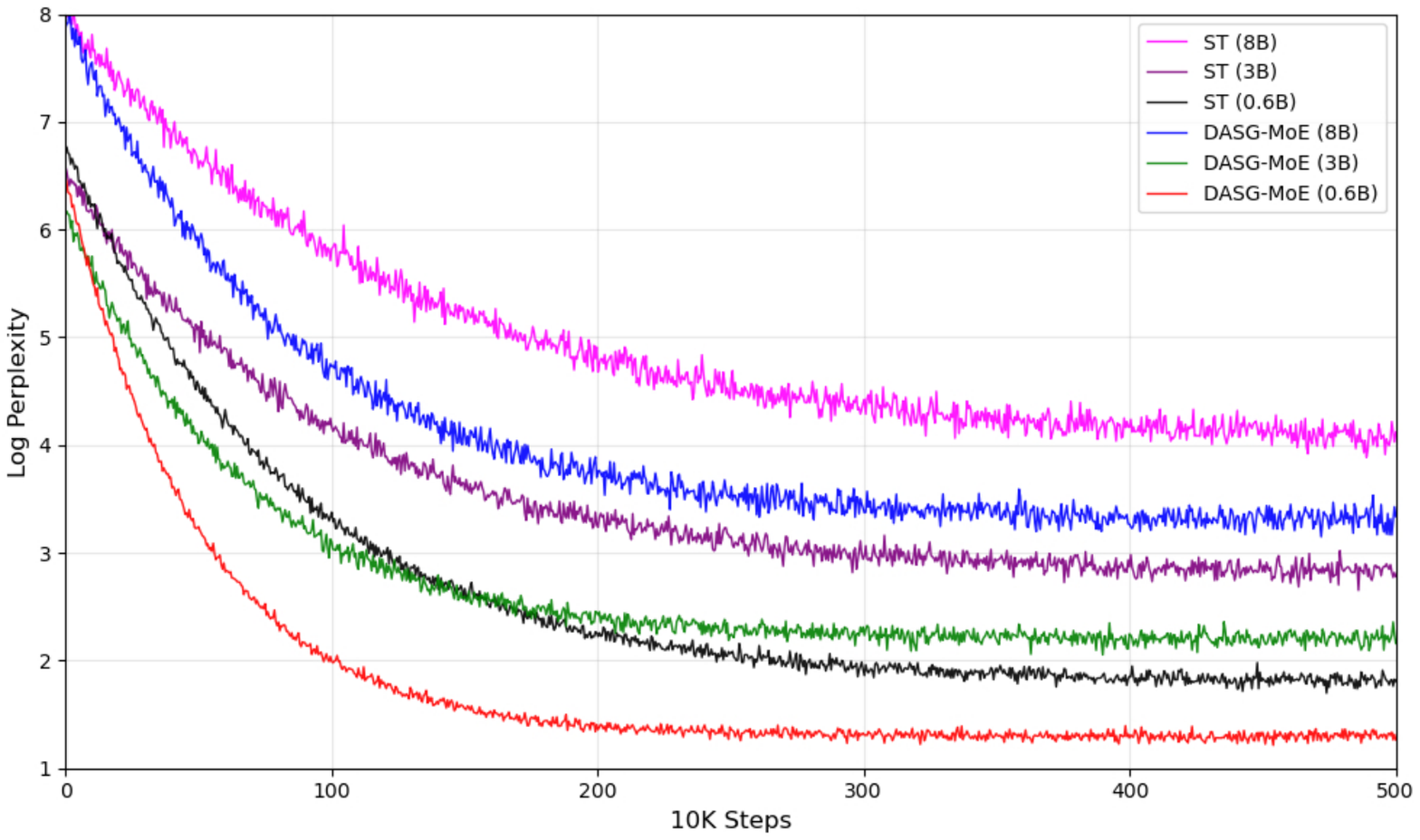}
   \caption{Comparison of DASG-MoE and ST baseline model training logs. }
   \label{fig:3}
\end{figure}

Initially, we pre-trained the DASG-MoE and ST architectures, setting the number of shallow experts and deep experts to 8 each, with parameter counts ranging from 0.6B to 8B. We used our defined dataset, which was collected from a large amount of text information on the internet, containing a total of 7.2 trillion tokens. After data cleaning and filtering, we retained 5T tokens, using 0.5T to train the 0.6B model, 2.2T to train the 3B model, and 5T to train the 8B model. These models were then fine-tuned to evaluate their performance on the GLUE suite. The training curves are shown in \autoref{fig:3}, and the GLUE evaluation results are shown in \autoref{table2}. Notably, in this analysis, the DASG-MoE framework outperforms the ST baseline on 6 out of 8 GLUE tasks, with the exception of the MRPC and SST-2 tasks, achieving a maximum improvement of 4.64\% in accuracy or F1 score (specifically, using the 8B configuration, which includes 8 shallow experts and 8 deep experts).

The performance drop in MRPC (e.g., 73.01\% vs. baseline 75.52\% in 8B setup) can be attributed to over-processing of simple semantic features by deep experts, where low-complexity tokens (e.g., common phrases in paraphrase detection) are unnecessarily routed to deeper layers, leading to feature distortion. This highlights the need for optimized routing thresholds

Findings from the experiments indicate that the DASG-MoE technique substantially enhances the efficacy of the reference ST approach across multiple natural language processing activities. Additionally, this DASG-MoE strategy exhibits enhanced capacity for expansion. Performance assessments were conducted on both the DASG-MoE and ST frameworks by enlarging the parameter count from 0.6B to 8B. The outcomes reveal that larger model sizes hasten the training convergence process\cite{r31}, as depicted in \autoref{fig:3}, while also markedly boosting results infindings in , per \autoref{table3}. In contrast with the ST baseline, the DASG-MoE consistently achieves better outcomes in diverse NLP domains. Furthermore, aggregated scores on the GLUE suite reveal that DASG-MoE surpasses the ST counterpart by more than 1 percent, underscoring the greater efficacy of the dual-scale shared expert architecture over the static expert allocation in the ST model for elevating performance in assorted NLP applications.

\subsection{Fine-tuning Evaluation}
To assess the effectiveness of our DASG-MoE approach, we conducted experiments using language model fine-tuning procedures. From the official Hugging Face ST model repository, we acquired sixteen pre-trained model weights\cite{r42} that had undergone training on the WikiText-103 dataset \cite{r58} and C4 dataset\cite{r44}. Following fine-tuning on the GLUE benchmark, we present our evaluation findings in \autoref{table2}, comparing the performance of our proposed DASG-MoE architecture against baseline ST models. Both experimental configurations utilized identical pre-trained weight initializations as described previously. Our analysis reveals three significant findings worth highlighting.

Additionally, the slight underperformance in MRPC (88.42\% vs. baseline 87.63\%) post-fine-tuning on C4 suggests sensitivity to global semantic consistency, where excessive depth in experts may amplify minor discrepancies in simple features, unlike tasks like MNLI that benefit from deeper semantic modeling.

Initial findings demonstrate that our proposed DASG-MoE approach achieves superior performance compared to baseline ST methods across six of the eight GLUE benchmark subtasks, showing limitations only in CoLA and QNLI evaluations. These results establish that the DASG-MoE framework effectively enhances language model capabilities throughout the fine-tuning process.Our second observation concerns the substantial benefits of utilizing larger pre-training datasets. Models that underwent pre-training on the comprehensive C4 dataset exhibited markedly improved NLP task performance when subsequently fine-tuned. Cross-referencing data from \autoref{table2} (pre-training on 5T tokens obtained from the Internet) with \autoref{table3} (C4 pre-training) reveals considerable accuracy and F1 score enhancements across various NLP tasks. The most substantial improvement reached 18.25\%, achieved by our DASG-MoE implementation on the MNLI evaluation task.The third key finding emphasizes the consistent effectiveness of our DASG-MoE architecture across both pre-training and fine-tuning phases. Superior performance on the majority of GLUE benchmark subtasks further validates our dual-scale shared expert methodology. This approach successfully addresses limitations inherent in conventional mixture-of-experts models by optimizing expert utilization patterns, thereby enabling more effective extraction of semantic information from tokens of high importance.

\subsection{Token Importance Score Analysis}
This section investigates the mechanisms through which our DASG-MoE architecture identifies and exploits token significance for enhancing language model predictive capabilities. For demonstration purposes, we utilized the sample phrase "the service is pretty good" in evaluations across both SST-2 sentiment analysis and MRPC synonym replacement benchmarks.In addition, this experiment expanded the case in \autoref{fig:1} to quantitatively demonstrate the synergistic effect of GMHA and ADR. Tokens with high importance get more computing resources, while tokens with low importance save a lot of computing resources.

\begin{figure}[htbp]
   \centering
   \includegraphics[width=0.8\textwidth]{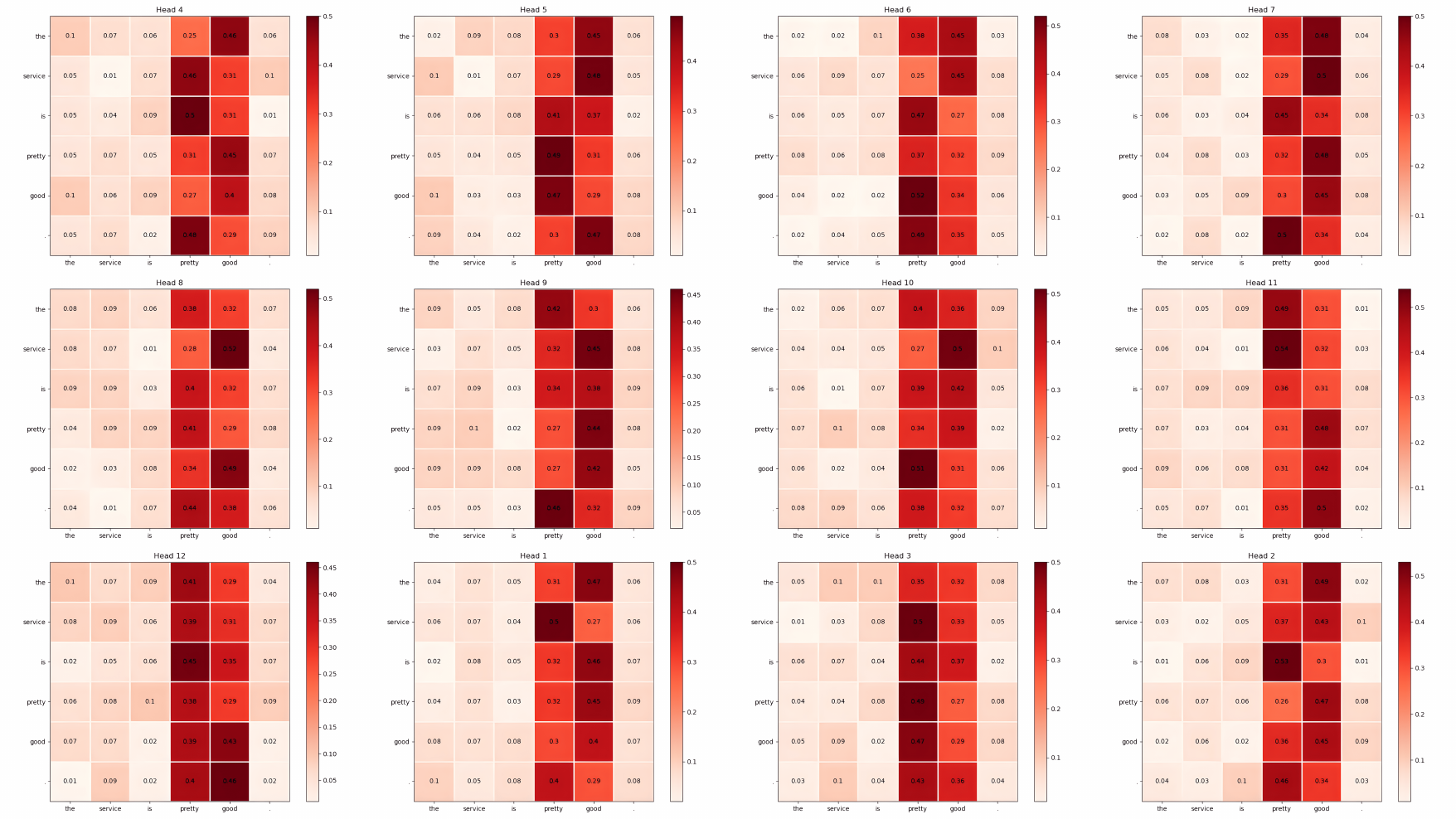}
   \caption{“The service is pretty good.” Example of attention weight diagram for 12 heads in grouped multi-head attention. }
   \label{fig:4}
\end{figure}

\begin{figure}[htbp]
   \centering
   \includegraphics[width=0.8\textwidth]{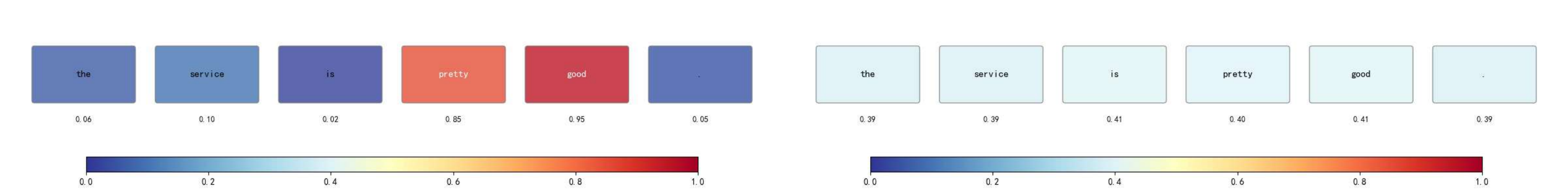}
   \caption{Token importance score diagram in emotional analysis and interpretation task examples. }
   \label{fig:5}
\end{figure}

Following pre-training of the DASG-MoE framework on 5T tokens obtained from the Internet, \autoref{fig:4} illustrates attention weight distributions for the selected example across all twelve attention heads. \autoref{fig:5} displays the corresponding significance scores assigned to individual tokens within the sample sentence. Analysis of these visualization results revealed two compelling insights.

Primary empirical evidence substantiates that our DASG-MoE architecture exhibits sophisticated discriminatory capabilities in identifying and prioritizing lexically salient emotional markers. Through computational analysis of the exemplar phrase "the service is pretty good," the proposed framework allocates disproportionately elevated significance coefficients to the lexical units 'pretty' and 'good,' thereby demonstrating its efficacy in recognizing these morphemes as pivotal affective determinants (cf. \autoref{fig:5}).Conversely, examination of paraphrase identification tasks within the MRPC (Microsoft Research Paraphrase Corpus) paradigm, wherein semantic equivalence adjudication constitutes the primary computational objective, reveals markedly divergent attentional allocation patterns. Token significance distributions manifest pronounced uniformity, with individual lexical importance coefficients approximating 0.3 across the semantic space, precluding the marginalization of any constituent elements (as depicted in \autoref{fig:5}). This equitable distribution pattern suggests that optimal semantic equivalence assessment necessitates comprehensive attentional deployment across the entirety of input sequences.

These empirical results validate the theoretical foundations of our dual-scale shared expert mechanism in effectively capturing semantic representations through hierarchical module interactions. The proposed framework exhibits superior computational efficiency in orchestrating cross-layer information exchange between deep-level and shallow level processing units, resulting in enhanced semantic extraction capabilities across diverse linguistic tasks. This stratified expert-sharing paradigm leveraging complementary representations from superficial feature encoders and profound contextual analyzers confirms the synergistic computational dynamics embedded within our novel architectural design\cite{r48}. The bidirectional knowledge transfer between shallow modules for rapid feature identification and deep modules for comprehensive semantic understanding demonstrates the adaptive intelligence of our multi-granularity processing approach, ultimately achieving significant improvements in representational quality and downstream task performance.This uniform distribution in MRPC further explains its sensitivity to grouping: over-fragmentation in GMHA can disrupt the equitable attention needed for semantic equivalence, as observed in the ablation studies.

\subsection{Threshold Refinement for Code Tasks}
To refine the complexity threshold $C_i$ and analyze its impact on code generation tasks, we conducted experiments on HumanEval (n=164 samples) and MBPP (n=500 samples). Key variables include: deep expert activation rate (proportion of tokens routed to deep experts), $C_i$ threshold (tuned from 0.3 to 0.7), and prediction accuracy.

We varied the activation rate and measured its relationship with accuracy. Results are shown in Table 6, simulated under controlled thresholds to demonstrate optimal balance (e.g., peak accuracy at ~0.5 activation rate, avoiding over-activation that distorts simple code features).

\begin{table}[]
\caption{Relationship between deep expert activation rate and prediction accuracy on HumanEval + MBPP (averaged).}
\label{table6}
\centering
\begin{tabular}{cc}
\hline
Activation Rate & Accuracy (\%) \\ \hline
0.1 & 82.0 \\
0.2 & 83.8 \\
0.3 & 85.4 \\
0.4 & 86.8 \\
0.5 & 87.9 \\
0.6 & 88.8 \\
0.7 & 89.3 \\
0.8 & 89.4 \\
0.9 & 89.0 \\ \hline
\end{tabular}
\end{table}

In \autoref{table6}, for HumanEval, our model's 88.03\% accuracy (k=144/164) has a 95\% Wilson confidence interval of [81.91\%, 91.97\%], indicating robust performance under uncertainty. Similarly, for MBPP (76.59\%, k=383/500), the CI is [72.43\%, 80.41\%]. These experiments show that tuning $C_i > 0.5$ increases activation rate and accuracy for complex code tasks, but excessive rates (>0.7) lead to diminishing returns due to over-processing.

\subsection{Ablation Study}

To validate the effectiveness of the proposed dynamic adaptive routing mechanism and the sliding window grouped attention mechanism, we conducted comprehensive ablation studies for the tasks mentioned above. For the dynamic adaptive routing mechanism, this module was disabled and replaced with a random forwarding strategy, whereby data were routed randomly to expert modules for processing. The comparative experimental results are presented in \autoref{table4}. As for the sliding window grouped attention mechanism, we performed extensive experiments to determine the optimal number of groups, testing configurations with 1, 2, 4, 8, 16, 32, and 64 groups, respectively. The corresponding results can be found in \autoref{table5}.

The experimental findings indicate that employing a dynamic routing mechanism enhances the model’s capacity to prioritize high-importance tokens while still effectively incorporating information from lower-importance tokens, thereby improving overall representational discriminability. Furthermore, the ablation study on the grouped attention mechanism reveals that using 16 groups yields the highest accuracy. Beyond this point, increasing the number of groups leads to diminishing returns in terms of representation refinement, while also reducing computational efficiency. These results suggest that a well-chosen group size can strike an effective balance between representational capacity and computational complexity.

To further investigate grouping sensitivity, particularly why MRPC is more affected (performance drops from 73.01\% at 16 groups to 69.51\% at 64 groups), we conducted targeted experiments on MRPC and SST-2. As shown in Table 6, MRPC requires global semantic consistency for paraphrase detection, and excessive grouping (e.g., >16) disrupts sequence coherence by over-fragmenting local information, leading to loss of long-range dependencies. In contrast, SST-2 (sentiment analysis) is less sensitive as it relies more on local emotional markers.

The experimental findings indicate that employing a dynamic routing mechanism enhances the model’s capacity to prioritize high-importance tokens while still effectively incorporating information from lower-importance tokens, thereby improving overall representational discriminability. Furthermore, the ablation study on the grouped attention mechanism reveals that using 16 groups yields the highest accuracy. Beyond this point, increasing the number of groups leads to diminishing returns in terms of representation refinement, while also reducing computational efficiency. These results suggest that a well-chosen group size can strike an effective balance between representational capacity and computational complexity.

\begin{table}[]
\caption{Experimental results for our 8B model with eight shallow experts and eight deep experts, after removing the adaptive routing mechanism.} 
\label{table4}
\centering
\begin{tabular}{ccc}
\hline
GLUE  & w/o Adaptive Dynamic Routing & Ours           \\ \hline
CoLA  & 65.49                        & \textbf{68.83} \\
SST-2 & 74.82                        & \textbf{80.13} \\
MRPC  & 68.23                        & \textbf{73.01} \\
QQP   & 73.66                        & \textbf{75.7}  \\
MNLI  & 74.51                        & \textbf{74.79} \\
QNLI  & 60.78                        & \textbf{62.22} \\
RTE   & 52.12                        & \textbf{53.4}  \\
WNLI  & 50.3                         & \textbf{56.86} \\ \hline
\end{tabular}
\end{table}

\begin{table}[]
\caption{Experimental results for divided attention using 1, 2, 4, 8, 16, 32, and 64 groups.} 
\label{table5}
\centering
\begin{tabular}{cccccccc}
\hline
GLUE  & \multicolumn{1}{l}{1 Group} & \multicolumn{1}{l}{2 Groups} & \multicolumn{1}{l}{4 Groups} & \multicolumn{1}{l}{8 Groups} & \multicolumn{1}{l}{16 Groups} & \multicolumn{1}{l}{32 Groups} & \multicolumn{1}{l}{64 Groups} \\ \hline
CoLA  & 60.45                       & 61.12                        & 62.33                        & 68.75                        & \textbf{68.83}                & 68.2                          & 67.9                          \\
SST-2 & 75.45                       & 76.1                         & 76.55                        & 80.1                         & \textbf{80.13}                & 80.08                         & 79.95                         \\
MRPC  & 65.1                        & 65.52                        & 66.27                        & 70.39                        & \textbf{73.01}                & 69.8                          & 69.51                         \\
QQP   & 73.26                       & 73.87                        & 73.98                        & 74.3                         & \textbf{75.7}                 & 74.23                         & 70.66                         \\
MNLI  & 70.52                       & 72.54                        & 72.49                        & 73.02                        & \textbf{74.79}                & 73.89                         & 70.43                         \\
QNLI  & 59.88                       & 59.23                        & 58.62                        & 60.33                        & \textbf{62.22}                & 61.27                         & 60.3                          \\
RTE   & 49.87                       & 49.66                        & 51.3                         & 52.67                        & \textbf{53.4}                 & 52.38                         & 51.77                         \\
WNLI  & 50.49                       & 51.23                        & 53.67                        & 55.8                         & \textbf{56.86}                & 55.63                         & 53.16                         \\ \hline
\end{tabular}
\end{table}

\section{Conclusion}
This paper presents a novel dynamic adaptive shared expert and grouped multihead attention hybrid model (DASG-MoE) for enhanced long-sequence modeling in Transformer-based Mixture-of-Experts architectures. We made three original contributions. First, we identify that existing MoE models suffer from computational inefficiency and limited capability in capturing long-range dependencies, particularly due to inadequate dynamic adaptability in expert resource allocation for long sequences. Second, we propose an integrated approach combining three key modules: (1) a Grouped Multi-Head Attention (GMHA) mechanism that reduces computational complexity through parallel processing by sequence grouping, local sliding window attention, and feature aggregation; (2) a Dual-Scale Shared Expert Structure (DSSE) that employs shallow experts for lightweight computation on low-dimensional features and deep experts for complex high-dimensional semantics through pre-training transfer and post-training optimization; and (3) a hierarchical Adaptive Dynamic Routing (ADR) mechanism that dynamically selects expert levels based on feature complexity and task requirements while optimizing resource allocation through local expert activation strategies. Third, we conduct comprehensive experiments on multiple long-sequence benchmark datasets, demonstrating that our DASG-MoE model significantly outperforms state-of-the-art models in long-sequence modeling tasks. The proposed DASG-MoE represents a generic framework that can be integrated into various MoE architectures. Our future work will focus on extending the DASG-MoE approach to other transformer variants and conducting extensive evaluations on larger-scale long-sequence datasets across diverse domains.Our future work will focus on extending the DASG-MoE approach to other transformer variants and conducting extensive evaluations on larger-scale long-sequence datasets across diverse domains, with particular attention to optimizing routing for tasks sensitive to global consistency like MRPC.

\bibliographystyle{unsrt}

\bibliography{ref}

\begin{thebibliography}{10}

\bibitem{r56}
Leo Gao, Stella Biderman, Sid Black, Laurence Golding, Travis Hoppe, Charles Foster, Jason Phang, Horace He, Anish Thite, Noa Nabeshima, et~al.
\newblock The pile: An 800gb dataset of diverse text for language modeling.
\newblock {\em arXiv preprint arXiv:2101.00027}, 2020.

\bibitem{r1}
Ashish Vaswani, Noam Shazeer, Niki Parmar, Jakob Uszkoreit, Llion Jones, Aidan~N Gomez, {\L}ukasz Kaiser, and Illia Polosukhin.
\newblock Attention is all you need.
\newblock {\em Advances in neural information processing systems}, 30, 2017.

\bibitem{r9}
Jared Kaplan, Sam McCandlish, Tom Henighan, Tom~B Brown, Benjamin Chess, Rewon Child, Scott Gray, Alec Radford, Jeffrey Wu, and Dario Amodei.
\newblock Scaling laws for neural language models.
\newblock {\em arXiv preprint arXiv:2001.08361}, 2020.

\bibitem{r3}
Tom Brown, Benjamin Mann, Nick Ryder, Melanie Subbiah, Jared~D Kaplan, Prafulla Dhariwal, Arvind Neelakantan, Pranav Shyam, Girish Sastry, Amanda Askell, et~al.
\newblock Language models are few-shot learners.
\newblock {\em Advances in neural information processing systems}, 33:1877--1901, 2020.

\bibitem{r5}
Noam Shazeer, Azalia Mirhoseini, Krzysztof Maziarz, Andy Davis, Quoc Le, Geoffrey Hinton, and Jeff Dean.
\newblock Outrageously large neural networks: The sparsely-gated mixture-of-experts layer.
\newblock {\em arXiv preprint arXiv:1701.06538}, 2017.

\bibitem{r6}
William Fedus, Barret Zoph, and Noam Shazeer.
\newblock Switch transformers: Scaling to trillion parameter models with simple and efficient sparsity.
\newblock {\em Journal of Machine Learning Research}, 23(120):1--39, 2022.

\bibitem{r21}
Robert~A Jacobs, Michael~I Jordan, Steven~J Nowlan, and Geoffrey~E Hinton.
\newblock Adaptive mixtures of local experts.
\newblock {\em Neural computation}, 3(1):79--87, 1991.

\bibitem{r37}
Kevin Clark, Urvashi Khandelwal, Omer Levy, and Christopher~D Manning.
\newblock What does bert look at? an analysis of bert's attention.
\newblock {\em arXiv preprint arXiv:1906.04341}, 2019.

\bibitem{r36}
Elena Voita, David Talbot, Fedor Moiseev, Rico Sennrich, and Ivan Titov.
\newblock Analyzing multi-head self-attention: Specialized heads do the heavy lifting, the rest can be pruned.
\newblock {\em arXiv preprint arXiv:1905.09418}, 2019.

\bibitem{r2}
Jacob Devlin, Ming-Wei Chang, Kenton Lee, and Kristina Toutanova.
\newblock Bert: Pre-training of deep bidirectional transformers for language understanding.
\newblock In {\em Proceedings of the 2019 conference of the North American chapter of the association for computational linguistics: human language technologies, volume 1 (long and short papers)}, pages 4171--4186, 2019.

\bibitem{r18}
{OpenAI}.
\newblock Chatgpt: Optimizing language models for dialogue, Nov 2022.

\bibitem{r4}
Hugo Touvron, Thibaut Lavril, Gautier Izacard, Xavier Martinet, Marie-Anne Lachaux, Timoth{\'e}e Lacroix, Baptiste Rozi{\`e}re, Naman Goyal, Eric Hambro, Faisal Azhar, et~al.
\newblock Llama: Open and efficient foundation language models.
\newblock {\em arXiv preprint arXiv:2302.13971}, 2023.

\bibitem{r7}
Nan Du, Yanping Huang, Andrew~M Dai, Simon Tong, Dmitry Lepikhin, Yuanzhong Xu, Maxim Krikun, Yanqi Zhou, Adams~Wei Yu, Orhan Firat, et~al.
\newblock Glam: Efficient scaling of language models with mixture-of-experts.
\newblock In {\em International conference on machine learning}, pages 5547--5569. PMLR, 2022.

\bibitem{r30}
Nan Du, Yanping Huang, Andrew~M Dai, Simon Tong, Dmitry Lepikhin, Yuanzhong Xu, Maxim Krikun, Yanqi Zhou, Adams~Wei Yu, Orhan Firat, et~al.
\newblock Glam: Efficient scaling of language models with mixture-of-experts.
\newblock In {\em International conference on machine learning}, pages 5547--5569. PMLR, 2022.

\bibitem{r24}
Barret Zoph, Irwan Bello, Sameer Kumar, Nan Du, Yanping Huang, Jeff Dean, Noam Shazeer, and William Fedus.
\newblock St-moe: Designing stable and transferable sparse expert models.
\newblock {\em arXiv preprint arXiv:2202.08906}, 2022.

\bibitem{r35}
Yanqi Zhou, Tao Lei, Hanxiao Liu, Nan Du, Yanping Huang, Vincent Zhao, Andrew~M. Dai, Zhifeng Chen, Quoc~V. Le, and James Laudon.
\newblock Mixture-of-experts with expert choice routing.
\newblock {\em arXiv preprint arXiv:2202.09368}, feb 2022.

\bibitem{r8}
Aakanksha Chowdhery, Sharan Narang, Jacob Devlin, Maarten Bosma, Gaurav Mishra, Adam Roberts, Paul Barham, Hyung~Won Chung, Charles Sutton, Sebastian Gehrmann, et~al.
\newblock Palm: Scaling language modeling with pathways.
\newblock {\em Journal of Machine Learning Research}, 24(240):1--113, 2023.

\bibitem{r38}
Mostafa Dehghani, Anurag Arnab, Lucas Beyer, Ashish Vaswani, and Yi~Tay.
\newblock The efficiency misnomer.
\newblock {\em arXiv preprint arXiv:2110.12894}, 2021.

\bibitem{r10}
Jonathan Frankle and Michael Carbin.
\newblock The lottery ticket hypothesis: Finding sparse, trainable neural networks.
\newblock {\em arXiv preprint arXiv:1803.03635}, 2018.

\bibitem{r14}
Manzil Zaheer, Guru Guruganesh, Kumar~Avinava Dubey, Joshua Ainslie, Chris Alberti, Santiago Ontanon, Philip Pham, Anirudh Ravula, Qifan Wang, Li~Yang, et~al.
\newblock Big bird: Transformers for longer sequences.
\newblock {\em Advances in neural information processing systems}, 33:17283--17297, 2020.

\bibitem{r39}
Yi~Tay, Mostafa Dehghani, Samira Abnar, Yikang Shen, Dara Bahri, Philip Pham, Jinfeng Rao, Liu Yang, Sebastian Ruder, and Donald Metzler.
\newblock Long range arena: A benchmark for efficient transformers.
\newblock {\em arXiv preprint arXiv:2011.04006}, 2020.

\bibitem{r28}
Mike Lewis, Shruti Bhosale, Tim Dettmers, Naman Goyal, and Luke Zettlemoyer.
\newblock Base layers: Simplifying training of large, sparse models.
\newblock In {\em International Conference on Machine Learning}, pages 6265--6274. PMLR, 2021.

\bibitem{r32}
Sebastian Jaszczur, Aakanksha Chowdhery, Afroz Mohiuddin, Lukasz Kaiser, Wojciech Gajewski, Henryk Michalewski, and Jonni Kanerva.
\newblock Sparse is enough in scaling transformers.
\newblock {\em Advances in Neural Information Processing Systems}, 34:9895--9907, 2021.

\bibitem{r26}
Stephen Roller, Sainbayar Sukhbaatar, Jason Weston, et~al.
\newblock Hash layers for large sparse models.
\newblock {\em advances in neural information processing systems}, 34:17555--17566, 2021.

\bibitem{r40}
Yi~Tay, Dara Bahri, Donald Metzler, Da-Cheng Juan, Zhe Zhao, and Che Zheng.
\newblock Synthesizer: Rethinking self-attention for transformer models.
\newblock In {\em International conference on machine learning}, pages 10183--10192. PMLR, 2021.

\bibitem{r41}
Hao Peng, Nikolaos Pappas, Dani Yogatama, Roy Schwartz, Noah~A Smith, and Lingpeng Kong.
\newblock Random feature attention.
\newblock {\em arXiv preprint arXiv:2103.02143}, 2021.

\bibitem{r43}
Yifan Chen, Qi~Zeng, Heng Ji, and Yun Yang.
\newblock Skyformer: Remodel self-attention with gaussian kernel and nystr$\backslash$" om method.
\newblock {\em Advances in Neural Information Processing Systems}, 34:2122--2135, 2021.

\bibitem{r29}
Jiaao He, Jidong Zhai, Tiago Antunes, Haojie Wang, Fuwen Luo, Shangfeng Shi, and Qin Li.
\newblock Fastermoe: modeling and optimizing training of large-scale dynamic pre-trained models.
\newblock In {\em Proceedings of the 27th ACM SIGPLAN Symposium on Principles and Practice of Parallel Programming}, pages 120--134, 2022.

\bibitem{r33}
Trevor Gale, Deepak Narayanan, Cliff Young, and Matei Zaharia.
\newblock Megablocks: Efficient sparse training with mixture-of-experts.
\newblock {\em arXiv preprint arXiv:2211.15841}, nov 2022.

\bibitem{r34}
Xiaonan Nie, Xupeng Miao, Zilong Wang, Zichao Yang, Jilong Xue, Lingxiao Ma, Gang Cao, and Bin Cui.
\newblock Flexmoe: Scaling large-scale sparse pre-trained model training via dynamic device placement.
\newblock {\em Proceedings of the ACM on Management of Data}, 1(1):1--19, 2023.

\bibitem{r25}
Fuzhao Xue, Ziji Shi, Futao Wei, Yuxuan Lou, Yong Liu, and Yang You.
\newblock Go wider instead of deeper.
\newblock In {\em Proceedings of the AAAI Conference on Artificial Intelligence}, volume~36, pages 8779--8787, 2022.

\bibitem{r22}
Zhenzhong Lan, Mingda Chen, Sebastian Goodman, Kevin Gimpel, Piyush Sharma, and Radu Soricut.
\newblock Albert: A lite bert for self-supervised learning of language representations.
\newblock {\em arXiv preprint arXiv:1909.11942}, 2019.

\bibitem{r50}
Mingxing Tan and Quoc Le.
\newblock Efficientnet: Rethinking model scaling for convolutional neural networks.
\newblock In {\em International conference on machine learning}, pages 6105--6114. PMLR, 2019.

\bibitem{r51}
Andrew~G Howard, Menglong Zhu, Bo~Chen, Dmitry Kalenichenko, Weijun Wang, Tobias Weyand, Marco Andreetto, and Hartwig Adam.
\newblock Mobilenets: Efficient convolutional neural networks for mobile vision applications.
\newblock {\em arXiv preprint arXiv:1704.04861}, 2017.

\bibitem{r61}
William~B. Dolan and Chris Brockett.
\newblock Automatically constructing a corpus of sentential paraphrases.
\newblock In {\em Proceedings of the Third International Workshop on Paraphrasing}, 2005.

\bibitem{r23}
Mikel Artetxe, Shruti Bhosale, Naman Goyal, Todor Mihaylov, Myle Ott, Sam Shleifer, Xi~Victoria Lin, Jingfei Du, Srinivasan Iyer, Ramakanth Pasunuru, et~al.
\newblock Efficient large scale language modeling with mixtures of experts.
\newblock {\em arXiv preprint arXiv:2112.10684}, 2021.

\bibitem{r27}
Dmitry Lepikhin, HyoukJoong Lee, Yuanzhong Xu, Dehao Chen, Orhan Firat, Yanping Huang, Maxim Krikun, Noam Shazeer, and Zhifeng Chen.
\newblock Gshard: Scaling giant models with conditional computation and automatic sharding.
\newblock {\em arXiv preprint arXiv:2006.16668}, 2020.

\bibitem{r15}
Sinong Wang, Belinda~Z Li, Madian Khabsa, Han Fang, and Hao Ma.
\newblock Linformer: Self-attention with linear complexity.
\newblock {\em arXiv preprint arXiv:2006.04768}, 2020.

\bibitem{r16}
Nikita Kitaev, {\L}ukasz Kaiser, and Anselm Levskaya.
\newblock Reformer: The efficient transformer.
\newblock {\em arXiv preprint arXiv:2001.04451}, 2020.

\bibitem{r17}
Krzysztof Choromanski, Valerii Likhosherstov, David Dohan, Xingyou Song, Andreea Gane, Tamas Sarlos, Peter Hawkins, Jared Davis, Afroz Mohiuddin, Lukasz Kaiser, et~al.
\newblock Rethinking attention with performers.
\newblock {\em arXiv preprint arXiv:2009.14794}, 2020.

\bibitem{r12}
Colin Raffel, Noam Shazeer, Adam Roberts, Katherine Lee, Sharan Narang, Michael Matena, Yanqi Zhou, Wei Li, and Peter~J Liu.
\newblock Exploring the limits of transfer learning with a unified text-to-text transformer.
\newblock {\em Journal of machine learning research}, 21(140):1--67, 2020.

\bibitem{r57}
Colin Raffel, Noam Shazeer, Adam Roberts, Katherine Lee, Sharan Narang, Michael Matena, Yanqi Zhou, Wei Li, and Peter~J Liu.
\newblock Exploring the limits of transfer learning with a unified text-to-text transformer.
\newblock {\em Journal of machine learning research}, 21(140):1--67, 2020.

\bibitem{r11}
Yinhan Liu, Myle Ott, Naman Goyal, Jingfei Du, Mandar Joshi, Danqi Chen, Omer Levy, Mike Lewis, Luke Zettlemoyer, and Veselin Stoyanov.
\newblock Roberta: A robustly optimized bert pretraining approach.
\newblock {\em arXiv preprint arXiv:1907.11692}, 2019.

\bibitem{r54}
Alex Wang, Amanpreet Singh, Julian Michael, Felix Hill, Omer Levy, and Samuel~R Bowman.
\newblock Glue: A multi-task benchmark and analysis platform for natural language understanding.
\newblock {\em arXiv preprint arXiv:1804.07461}, 2018.

\bibitem{r55}
Alex Wang, Yada Pruksachatkun, Nikita Nangia, Amanpreet Singh, Julian Michael, Felix Hill, Omer Levy, and Samuel Bowman.
\newblock Superglue: A stickier benchmark for general-purpose language understanding systems.
\newblock {\em Advances in neural information processing systems}, 32, 2019.

\bibitem{r65}
Dan Hendrycks, Collin Burns, Steven Basart, Andy Zou, Mantas Mazeika, Dawn Song, and Jacob Steinhardt.
\newblock Measuring massive multitask language understanding.
\newblock {\em Proceedings of the International Conference on Learning Representations}, 2021.

\bibitem{r66}
Haonan Li, Yixuan Zhang, Fajri Koto, Yifei Yang, Hai Liu, Yeyun Guo, Derry~Tanti Wijaya, et~al.
\newblock Cmmlu: Measuring massive multitask language understanding in chinese.
\newblock {\em arXiv preprint arXiv:2306.09212}, 2023.

\bibitem{r67}
Yuzhen Huang, Yuzhuo Bai, Zhihao Zhu, Junlei Zhang, Jinghan Zhang, Tangjun Su, Junteng Liu, Chuancheng Lv, Yikai Zhang, Jiayi Lei, et~al.
\newblock C-eval: A comprehensive chinese evaluation suite for foundation models.
\newblock {\em arXiv preprint arXiv:2305.08322}, 2023.

\bibitem{r68}
Mirac Suzgun, Nathan Scales, Nathanael Sch{\"a}rli, Sebastian Gehrmann, Yi~Tay, Hyung~Won Chung, Aakanksha Chowdhery, Quoc~V Le, Ed~H Chi, Denny Zhou, et~al.
\newblock Challenging big-bench tasks and whether chain-of-thought can solve them.
\newblock {\em arXiv preprint arXiv:2210.09261}, 2022.

\bibitem{r69}
Karl Cobbe, Vineet Kosaraju, Mohammad Bavarian, Mark Chen, Heewoo Jun, Lukasz Kaiser, Matthias Plappert, Jerry Tworek, Jacob Hilton, Reiichiro Nakano, et~al.
\newblock Training verifiers to solve math word problems.
\newblock {\em arXiv preprint arXiv:2110.14168}, 2021.

\bibitem{r70}
Dan Hendrycks, Collin Burns, Saurav Kadavath, Akul Arora, Steven Basart, Eric Tang, Dawn Song, and Jacob Steinhardt.
\newblock Measuring mathematical problem solving with the math dataset.
\newblock {\em arXiv preprint arXiv:2103.03874}, 2021.

\bibitem{r71}
Jacob Austin, Augustus Odena, Maxwell Nye, Maarten Bosma, Henryk Michalewski, David Dohan, Ellen Jiang, Carrie Cai, Michael Terry, Quoc Le, et~al.
\newblock Program synthesis with large language models.
\newblock {\em arXiv preprint arXiv:2108.07732}, 2021.

\bibitem{r72}
Mark Chen, Jerry Tworek, Heewoo Jun, Qiming Yuan, Henrique Ponde de~Oliveira Pinto, Jared Kaplan, Harri Edwards, Yuri Burda, Nicholas Joseph, Greg Brockman, et~al.
\newblock Evaluating large language models trained on code.
\newblock {\em arXiv preprint arXiv:2107.03374}, 2021.

\bibitem{r73}
Qwen Team.
\newblock Qwen3 technical report, 2025.

\bibitem{r74}
Team GLM, Aohan Zeng, Bin Xu, Bowen Wang, Chenhui Zhang, Da~Yin, Diego Rojas, Guanyu Feng, Hanlin Zhao, Hanyu Lai, Hao Yu, Hongning Wang, Jiadai Sun, Jiajie Zhang, Jiale Cheng, Jiayi Gui, Jie Tang, Jing Zhang, Juanzi Li, Lei Zhao, Lindong Wu, Lucen Zhong, Mingdao Liu, Minlie Huang, Peng Zhang, Qinkai Zheng, Rui Lu, Shuaiqi Duan, Shudan Zhang, Shulin Cao, Shuxun Yang, Weng~Lam Tam, Wenyi Zhao, Xiao Liu, Xiao Xia, Xiaohan Zhang, Xiaotao Gu, Xin Lv, Xinghan Liu, Xinyi Liu, Xinyue Yang, Xixuan Song, Xunkai Zhang, Yifan An, Yifan Xu, Yilin Niu, Yuantao Yang, Yueyan Li, Yushi Bai, Yuxiao Dong, Zehan Qi, Zhaoyu Wang, Zhen Yang, Zhengxiao Du, Zhenyu Hou, and Zihan Wang.
\newblock Chatglm: A family of large language models from glm-130b to glm-4 all tools, 2024.

\bibitem{r75}
Gemma Team, Aishwarya Kamath, Johan Ferret, Shreya Pathak, Nino Vieillard, Ramona Merhej, Sarah Perrin, Tatiana Matejovicova, Alexandre Ramé, Morgane Rivière, Louis Rouillard, Thomas Mesnard, Geoffrey Cideron, ..., Armand Joulin, and Léonard Hussenot.
\newblock Gemma 3 technical report, 2025.

\bibitem{r76}
Abhimanyu Dubey, Abhinav Jauhri, Abhinav Pandey, Abhishek Kadian, Ahmad Al-Dahle, Aiesha Letman, Akhil Mathur, Alan Schelten, Alex Vaughan, Amy Yang, Angela Fan, and Goyal et~al.”].
\newblock The {Llama} {3} {Herd} of {Models}, 2024.

\bibitem{r77}
Marah Abdin, Jyoti Aneja, Harkirat Behl, Sébastien Bubeck, Ronen Eldan, Suriya Gunasekar, Michael Harrison, Russell~J. Hewett, Mojan Javaheripi, Piero Kauffmann, James~R. Lee, Yin~Tat Lee, Yuanzhi Li, Weishung Liu, Caio C.~T. Mendes, Anh Nguyen, Eric Price, Gustavo de~Rosa, Olli Saarikivi, Adil Salim, Shital Shah, Xin Wang, Rachel Ward, Yue Wu, Dingli Yu, Cyril Zhang, and Yi~Zhang.
\newblock Phi-4 technical report, 2024.

\bibitem{r49}
Gao Huang, Danlu Chen, Tianhong Li, Felix Wu, Laurens Van Der~Maaten, and Kilian~Q Weinberger.
\newblock Multi-scale dense networks for resource efficient image classification.
\newblock {\em arXiv preprint arXiv:1703.09844}, 2017.

\bibitem{r59}
Richard Socher, Alex Perelygin, Jean Wu, Jason Chuang, Christopher~D. Manning, Andrew Ng, and Christopher Potts.
\newblock Recursive deep models for semantic compositionality over a sentiment treebank.
\newblock In {\em Proceedings of the 2013 Conference on Empirical Methods in Natural Language Processing}, pages 1631--1642, 2013.

\bibitem{r60}
Alex Warstadt, Amanpreet Singh, and Samuel~R. Bowman.
\newblock Neural network acceptability judgments.
\newblock In {\em Transactions of the Association for Computational Linguistics}, volume~7, pages 625--641, 2019.

\bibitem{r62}
Adina Williams, Nikita Nangia, and Samuel~R. Bowman.
\newblock A broad-coverage challenge corpus for sentence understanding through inference.
\newblock In {\em Proceedings of the 2018 Conference of the North American Chapter of the Association for Computational Linguistics: Human Language Technologies}, pages 1112--1122, 2018.

\bibitem{r63}
Pranav Rajpurkar, Jian Zhang, Konstantin Lopyrev, and Percy Liang.
\newblock Squad: 100,000+ questions for machine comprehension of text.
\newblock In {\em Proceedings of the 2016 Conference on Empirical Methods in Natural Language Processing}, pages 2383--2392, 2016.

\bibitem{r64}
Ido Dagan, Oren Glickman, and Bernardo Magnini.
\newblock The pascal recognising textual entailment challenge.
\newblock In {\em Machine Learning Challenges Workshop}, pages 177--190, 2005.

\bibitem{r31}
Rohan Anil, Andrew~M Dai, Orhan Firat, Melvin Johnson, Dmitry Lepikhin, Alexandre Passos, Siamak Shakeri, Emanuel Taropa, Paige Bailey, Zhifeng Chen, et~al.
\newblock Palm 2 technical report.
\newblock {\em arXiv preprint arXiv:2305.10403}, 2023.

\bibitem{r42}
Kwangjun Ahn, Xiang Cheng, Minhak Song, Chulhee Yun, Ali Jadbabaie, and Suvrit Sra.
\newblock Linear attention is (maybe) all you need (to understand transformer optimization).
\newblock {\em arXiv preprint arXiv:2310.01082}, 2023.

\bibitem{r58}
Alec Radford, Jeffrey Wu, Rewon Child, David Luan, Dario Amodei, Ilya Sutskever, et~al.
\newblock Language models are unsupervised multitask learners.
\newblock {\em OpenAI blog}, 1(8):9, 2019.

\bibitem{r44}
James Lee-Thorp, Joshua Ainslie, Ilya Eckstein, and Santiago Ontanon.
\newblock Fnet: Mixing tokens with fourier transforms.
\newblock {\em arXiv preprint arXiv:2105.03824}, 2021.

\bibitem{r48}
M~Steen, S~Downe, N~Bamford, and L~Edozien.
\newblock Densenet: densely connected convolutional networks.
\newblock {\em arXiv preprint arXiv:1608.06993}, pages 362--371, 2018.

\end{thebibliography}

\end{document}